\documentclass[conference]{IEEEtran}
\IEEEoverridecommandlockouts

\usepackage{cite}
\usepackage{amsmath,amssymb,amsfonts}
\usepackage{algorithmic}
\usepackage{graphicx}
\usepackage{textcomp}
\usepackage{xcolor}
\usepackage{multirow}
\usepackage{tikz}
\usepackage{pgfplots}
\usepackage{float}
\usepackage{threeparttable}
\usetikzlibrary{patterns}
\usepackage{regexpatch,fancyvrb,xparse}
\usepackage{hyperref}
\usepackage{caption}
\usepackage{authblk}

\def\BibTeX{{\rm B\kern-.05em{\sc i\kern-.025em b}\kern-.08em
    T\kern-.1667em\lower.7ex\hbox{E}\kern-.125emX}}

\IEEEoverridecommandlockouts\IEEEpubid{\makebox[\columnwidth]{978-1-6654-0435-8/21/\$31.00~\copyright~2021~IEEE \hfill}\hspace{\columnsep}\makebox[\columnwidth]{ }}

\begin{document}

\title{Automatically Detecting Cyberbullying Comments on Online Game Forums}

\author[1,2,*]{Hanh Hong-Phuc Vo}
\author[1,2,*]{Hieu Trung Tran}
\author[1,2,†]{Son T. Luu}
\affil[1]{University of Information Technology, Ho Chi Minh City, Vietnam}
\affil[2]{Vietnam National University Ho Chi Minh City, Vietnam}

\affil[ ]{Email: *\textit {\{18520275,18520754\}@gm.uit.edu.vn}, †\textit{sonlt@uit.edu.vn}}
\IEEEoverridecommandlockouts
\IEEEpubid{\makebox[\columnwidth]{978-1-6654-0435-8/21/\$31.00~\copyright2021 IEEE \hfill} \hspace{\columnsep}\makebox[\columnwidth]{ }}
\maketitle
\IEEEpubidadjcol
\begin{abstract}
Online game forums are popular to most of game players. They use it to communicate and discuss the strategy of the game, or even to make friends. However, game forums also contain abusive and harassment speech, disturbing and threatening players. Therefore, it is necessary to automatically detect and remove cyberbullying comments to keep the game forum clean and friendly. We use the Cyberbullying dataset collected from World of Warcraft (WoW) and League of Legends (LoL) forums and train classification models to automatically detect whether a comment of a player is abusive or not. The result obtains 82.69\% of macro F1-score for LoL forum and 83.86\% of macro F1-score for WoW forum by the Toxic-BERT model on the Cyberbullying dataset.
\end{abstract}


\section{Introduction}
\label{intro}
Currently, offensive comments are common issues on gaming forums. These comments are often negative, harmful, and abusive, causing a pessimistic impression on the gaming community. To prevent offensive comments by users on forums, administrators have to manually find each comment, then hide it or ban the user whose comments are abusive. However, this approach seems impossible on big forums with huge user comments. Therefore, automatically detecting offensive comments on game forums is necessary to help administrators find and hide negative comments or even ban users who post abusive and harassment comments. We solve the problem of automatically detecting cyberbullying comments on the forum of League of Legends and World of Warcraft, which are popular to most game players.

We use machine learning approaches for text classification tasks to automatically classify whether a comment from a game player is abusive or not. We also propose alternative solutions such as data pre-processing and data enhancement to increase the performance of classifiers. Then, we evaluate the empirical results to find the best models on the Cyberbullying dataset. Our approaches to machine learning include: traditional machine learning models (Logistic Regression, SVM), deep neural models (Text-CNN, GRU), and transfer learning models (Toxic BERT). Finally, based on the empirical results, we choose the highest performance model to build a tool for helping administrators detect and hide abusive comments on online game forums.

The paper is organized as follows. Section 1 introduces the topic and motivation to study the aforementioned task. Section 2 presents related work on the problem of identifying abusive comments. Section 3 describes the Cyberbullying dataset that is used for training machine learning models. Section 4 and section 5 illustrate our approaches and empirical results. Finally, Section 6 concludes our work.
\section{Related work}
\label{related_works}
In 2017, Wulczyn et al. analyzed Wikipedia comments to find personally offensive comments with a dataset of over 100,000 high-quality human-labeled comments and 63M machine-labeled ones \cite{wulczyn2017ex}. The dataset achieved the best AUG result (96.59\%) using multi-layer perceptrons (MLP) model on ED $($empirical distribution$)$ label with char n-gram. In the case of a model trained on ED labels, the attack score represents the predicted fraction of annotators who would consider the comment an attack. 

Also in 2017, Davidson et al. conducted the study on Automated Hate Speech Detection and the Problem of Offensive Language \cite{davidson2017automated}. The authors constructed a dataset from tweets on Twitter, then trained them  with machine learning models such as Logistic Regression, Naive Bayes, Decision Trees, Random Forests, and Linear SVMs. After trained, the best performance model, which is Logistic Regression, had the results of 91.00\% by Precision and 90.00\% by Recall and F1-scores.

In addition, in 2018, Ribeiro et al. performed characterizing and detecting hateful users on Twitter \cite{ribeiro2018characterizing}. Comments in the dataset are collected from Twitter. However, unlike other datasets, this dataset focuses on the user, not the content of the user comments. The author crawled the dataset from 100,386 users with up to 200 tweets per user. They then selected a subsample to be annotated. There are 4,972 comments manually annotated users, of which 544 users labeled as hateful. They also implemented dataset models such as GraphSage, AdaBoost, GradBoost. Using both user's feature and glove properties together with the GraphSage model gives the best results. Additionally, AdaBoos delivers a good AUC score, but mis-classifying many of the normal users is negative, resulting in low accuracy and F1-score.

Besides, many multilingual datasets for abusive comment recognition are built in different languages, such as hatEval dataset from SemEval-2019 Task 5 contest (2019) \cite{basile-etal-2019-semeval} on identifying offensive comments about women and migrants in English and Spanish, and CONAN dataset is used to identify comments inciting violence in the English, French and Italian languages \cite {chung-etal-2019-conan}.
\section{Dataset}
\label{dataset}
The Cyberbullying dataset\footnote{\url{http://ub-web.de/research/}} \cite{bretschneider2016detecting} includes two parts: comments of players on League of Legends (LoL) and World of Warcraft (WoW) forums. The data from League of Legends forum contains 17,354 comments with 207 offensive cases, and data from the World of Warcraft forum contains 16,975 comments with 137 offensive cases. All comments are annotated manually and reviewed by experts. The dataset contains English comments on 20 different topics. There are two labels in the dataset, label 1 indicates offensive comments, and label 0 indicates non-offensive comments. Table \ref{tab:ti_le_nhan} show the data distribution of labels on two forums - LoL and WoW.

\begin{table}[H]
    \caption{Samples in the Cyberbullying dataset.}
    \label{tab:vi_du_trong_bo_du_lieu}
    \begin{tabular}{|p{0.3cm}|p{6.6cm}|p{0.6cm}|}
        \hline
        \textbf{No.} & \centering \textbf{Comment} &\textbf{Label} \\
        \hline
        \multicolumn{3}{|c|}{\textbf{Comments on League of Legends forum}} \\
        \hline
        1 &  We can only hope. & 0 \\
        \hline
        2 &  This is why we don't account share children. & 0 \\
        \hline
        3 &  This is from Season Stay down bitch fanboy loser. & 1 \\
        \hline
        4 & Rest in pieces. fk u & 1 \\
        \hline
        \multicolumn{3}{|c|}{\textbf{Comments on World of Warcraft forum}} \\
        \hline
        1 & This thread is depressing, progress is slow. & 0 \\
        \hline
        2 &  But they could give us free transfers in the meantime. & 0 \\
        \hline
        3 &  I didnt nuked it. I ran the all road and can tag him at k hp. Im a druid and i used cyclone time. So dont swear to me you idiot & 1 \\
        \hline
        4 & when you troll say a good time its . you down it at . Braindead & 1 \\
        \hline
    \end{tabular}
\end{table}

In addition, according to Table \ref{tab:vi_du_trong_bo_du_lieu}, comments are often written in an informal form, using a lot of slang, abbreviations, strange characters and conversations centered on the exchanges between players on game forums. In addition, it can be seen that comments from the Cyberbullying dataset often use informal words such as abbreviations and slang, for example: “u”, and “fk”. This is a common feature for comments on social media platforms \cite{fan2014power}. 

\begin{table}[H]
    \centering
    \caption{Distribution of labels by percentage on two forums in the Cyberbullying dataset.}
    \label{tab:ti_le_nhan}
    \begin{tabular}{|c|c|c|c|}
        \hline
        \textbf{Forum} & \textbf{Label 0} & \textbf{Label 1} \\
        \hline
        League of Legends  & 17,354 (98.81\%) & 207 (1.19\%) \\
        \hline
        World of Warcraft & 16,975 (99.19\%) & 137 (0.81\%) \\
        \hline         
    \end{tabular}
\end{table}

Furthermore, as showed in Table \ref{tab:ti_le_nhan}, the distribution of labels is imbalanced on both LoL and WoW forums in the dataset. Most of the comments are skewed to label 0, which account for 98.81\% on the LoL forum and 99.19\% on the WoW forum, respectively. This indicates that Cyberbullying dataset is imbalanced, and most of the comments are non-offensive comments. 

\begin{table}[H]
    \centering
    \caption{Average length of comments in the Cyberbullying dataset.}
    \label{tab:cmtlength}
    \begin{tabular}{|c|c|c|}
        \hline
        \textbf{Forum} & \textbf{Label 0} & \textbf{Label 1} \\
        \hline
        League of Legends & 102.95 & 137.60 \\
        \hline
        World of Warcraft  & 105.42 & 165.58\\
        \hline         
    \end{tabular}
\end{table}

Finally, Table \ref{tab:cmtlength} shows the average length of comments in the Cyberbullying dataset. The average length of comments is computed by word level.  According to Table \ref{tab:cmtlength}, the average length of comments on label 0 and label 1 are 102.95 and 137.60 respectively on the LoL forum, and 105.42 and 165.68 respectively on the WoW forum. In general, the average length of comments on label 1 are longer than the average length of comments on label 0 for both two forums.

\section{Methodologies}
\label{methodologies}
\subsection{Text classification task}
Text classification is one of the most common tasks in machine learning and natural language processing. Aggarwal and Zhai (2012) defined the text classification task as given the training set D = $\{X_1, ..., X_n \}$, where each data point $X_i \in D$ is labeled with a value derived from the set of labels are numbered $\{1…k\}$. The training data is built into a classification model. With the test set given, the classification model predicts the label for each data point in the test set \cite{aggarwal2012survey}.


\begin{figure}[!h]
    \centering
    \includegraphics[scale=0.45]{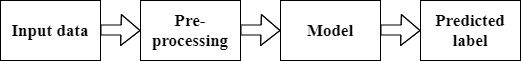}
    \caption{Model overview for detecting cyberbullying comments.}
    \label{fig_model_overview}
\end{figure}

Figure \ref{fig_model_overview} describes our model overview for detecting cyberbullying comments. With input comments, the model does pre-processing steps including splitting comments into tokens, removing unimportant characters, using SMOTE to generate new examples and using word embedding (GloVe, fastText) to present vectors for input data. Next, the pre-processed data are fit into classification models including traditional machine learning models, deep neural models, and transformer learning model. Finally, the output is labels that was predicted by classification models.

\subsection{SMOTE}
Synthetic Minority Oversampling Technique (SMOTE) is a technique to increase the number of samples for the minority class. It is widely used to handle imbalanced datasets. SMOTE works by selecting data points in the feature space and drawing a line between those points and the nearest neighbors. Then based on this line to select new data for the minority class. New data points are generated from the original data points to increase the amount of minority class. This is an effective method for solving imbalance in datasets \cite{chawla2002smote}.

\subsection{Word embedding}
Word embedding is a method of word representation that used in natural language processing. This method represents a word (or phrase) by a real number vector that was mapped from a corpus of text. Using word embedding improves the performance of the predictive model \cite{character2018end}. 

In this paper, we use sets of word representation vectors such as GloVe \cite {pennington2014glove} and fastText \cite{grave-etal-2018-learning}. The glove.6B.300d word embedding\footnote{\url{https://nlp.stanford.edu/projects/glove/}} comprises 300 dimensions vectors over 400,000 vocabularies, which is built on Wikipedia and Gigaword corpora. The fastText \footnote {\url{https://fasttext.cc/docs/en/crawl-vectors.html}} is trained on Common Crawl and Wikipedia datasets using CBOW with position-weight in 300 dimensions with 5-grams features.

\subsection{Traditional machine learning models}
\textbf{Logistic Regression} is one of the basic classification models commonly used for binary classification problems. For classification problems, we have to convert texts to vectors before fitting them into the training model\cite{nigam1999using}.

\textbf{Support Vector Machine (SVM)} is a classification algorithm based on the concept of hyper-planes, which divide data into different classes. This algorithm is commonly used for classification and regression problems. Similar to Logistic Regression, SVM converts text data into vector form before fitting into the training model \cite{chang2011libsvm}.

\subsection{Deep neural models}
\textbf{Text-CNN} is a multi-layered neural network architecture developed for the classification problem. CNN model uses Convolutional layers (CONV layer) to extract valuable information from data \cite{kim-2014-convolutional}.

\textbf{GRU} is an improved version of the standard recurrent neural network (RNN). To solve the vanishing gradient problem of a standard RNN, GRU uses special gates including update gate and reset gate. Basically, these are two vectors which decide what information should be passed to the output. The special thing about them is that they can be trained to keep information from long ago, without washing it through time or remove information which is relevant to the prediction\cite{cho-etal-2014-learning}.

\subsection{Transformer model}
BERT is a transformer architecture created by Devlin et al.\cite{devlin2018bert} and become popular to many of natural language processing tasks. With pre-trained models on large and unsupervised data, BERT can be fine-tuned later in specific downstream tasks such as text classification and question answering. Toxic-BERT (also called as Detoxify) \cite{Detoxify} is a model based on BERT architecture that is used for toxic comments classification task. Toxic-BERT is trained on three different toxic dataset comes from three Jigsaw challenges. We implement the Toxic-BERT model on the Cyberbullying dataset for detecting cyberbullying comments from players.
\section{Empirical results}
\label{empirical}
\subsection{Model parameters}
We implemented the Logistic Regression and SVM with random\_state equal to 0, C equal to 1, and max\_feature equal to 13,000. In addition, we implemeneted the Text-CNN and GRU with 5 epochs, batch\_size equal to 512, sequence\_length equal to 300, conv\_layer\_size equal to 5, 128 units, dropout equal to 0.1, and using sigmoid activation function. Finally, we implement the Toxic-BERT with 5 epochs, train\_batch\_size equal to 16, and test\_batch\_size equal to 8. We used the Simple transformer\footnote{\url{https://simpletransformers.ai/}} for implementing the Toxic-BERT model.





\subsection{Data preparation}
Before running the model, we do pre-processing steps as follows: (1) we remove special characters in the comments (except for the "*") and keep only the letters. The characters "*"\: represent encoded offensive words. These are handled by changing to the word "beep", (2) we split comments into tokens by using the TweetTokenizer of NLTK library, (3) we converted comments to lowercase, and (4) we remove stop words like "the", "in", "a", "an" because they have less meaning in the sentence.

We use the k-fold cross-validation with 5 folds (k = 5) on the Cyberbullying dataset. The dataset is randomly divided into 5 equal parts with proportion 8:2 for train set and test set respectively. The final result is the mean of 5 folds. 

\subsection{Empirical results}
Table \ref{tab:ket_qua_thuc_nghiem} presents the results of the models on two forums, LoL and WoW of the dataset. According to Table \ref{tab:ket_qua_thuc_nghiem}, the results of the models on the two forums are similar.

For traditional machine learning models, the macro F1-score and accuracy are quite high disparities. The reason is the imbalanced distribution of labels in the dataset as described in Section \ref{dataset}.

For deep neural models, the Text-CNN model with the GloVe word embedding gives the best results by macro F1 score, which are 80.68\% on the LoL forum and  83.10\% on the WoW forum, respectively. Besides, there is a discrepancy between Accuracy and macro F1 scores on deep neural models due to unbalanced data. However, this difference is not as much as traditional machine learning models.

Among the models, Toxic-BERT gives the highest results according to the macro F1-score on both forums, which are 82.69\% on LoL forum and 83.86\% on the WoW forum, respectively according to Table \ref{tab:ket_qua_thuc_nghiem}.

To handle the imbalanced in the dataset, we apply the SMOTE method to increasing the data in minor label. After using the SMOTE method, on traditional machine learning models, the results by macro F1-score  increased significantly. For Logistic Regression, the macro F1-score increases 10.49\% and 11.41\% on the LoL forum and WoW forums, respectively after using SMOTE. Same as Logistic Regression, the results by macro F1-score on the SVM increases 8.95\% and 11.41\% on the LoL forum and WoW forums, respectively after using SMOTE. However, SMOTE does not augments the performance of deep neural models on both forums.

\begin{table}[H]
    \centering
    \caption{Empirical results of the Cyberbullying dataset.}
    \begin{tabular}{|c|c|c|c|c|}
    \hline
    \multicolumn{5}{|c|}{\textbf{League of Legends Forum}} \\
    \hline
    \multirow{2}{*}{Model} & \multicolumn{2}{|c|}{SMOTE} & \multicolumn{2}{|c|}{NOT SMOTE} \\
    \cline{2-5}
    \textbf{} & \textbf{Acc}(\%) & \textbf{F1-sc(\%) } & \textbf{Acc}(\%) & \textbf{F1-sc(\%) }\\
    \hline
    Logistic Regression & 97.02 & 60.19 & 98.81 & 49.70\\
    \hline
    SVM & 98.20 & 58.65 & 98.81 & 49.70 \\
    \hline
    Text-CNN + fastText & 97.87 & 75.30 & 99.29 & 75.67\\ 
    \hline
    Text-CNN+GloVe & 97.06 & 73.19 & 99.37 & 80.68\\ 
    \hline
    GRU + fastText & 96.69 & 66.37 & 99.18 & 78.46\\ 
    \hline
    GRU + GloVe & 95.54 & 66.50 & 99.32 & 75.53\\ 
    \hline
    \textbf{Toxic-BERT} & - & - & \textbf{99.38} & \textbf{82.69} \\
    \hline
    \multicolumn{5}{|c|}{\textbf{ World of Warcraft Forum}} \\
    \hline
    \multirow{2}{*}{Model} & \multicolumn{2}{|c|}{SMOTE} & \multicolumn{2}{|c|}{NOT SMOTE} \\
    \cline{2-5}
    \textbf{} & \textbf{Acc}(\%) & \textbf{F1-sc(\%) } & \textbf{Acc}(\%) & \textbf{F1-sc(\%) }\\
    \hline
    Logistic Regression & 98.11 & 63.50 & 99.19 & 49.79\\
    \hline
    SVM & 98.74 & 61.20 & 99.19 & 49.79 \\
    \hline
    Text-CNN + fastText & 97.43 & 73.07 & 99.48 & 75.37 \\ 
    \hline
    Text-CNN+GloVe & 97.56 & 74.40 & 99.62 & 83.10\\ 
    \hline
    GRU + fastText & 96.27 & 66.19 & 99.49 & 74.44 \\
    \hline
    GRU + GloVe & 97.77 & 71.14 & 99.47 & 72.99 \\ 
    \hline
    \textbf{Toxic-BERT} & - & - & \textbf{99.58} & \textbf{83.86} \\
    \hline
    \end{tabular}
    \label{tab:ket_qua_thuc_nghiem}
\end{table}

\subsection{Error analysis}
Since the results of the models on the two forums are similar according to Table \ref{tab:ket_qua_thuc_nghiem}, we analyze the error predictions based on the LoL forum in the Cyberbullying dataset. 

\begin{figure}[H]
    \begin{minipage}{0.21\textwidth}
        \centering
        \includegraphics[scale=0.33]{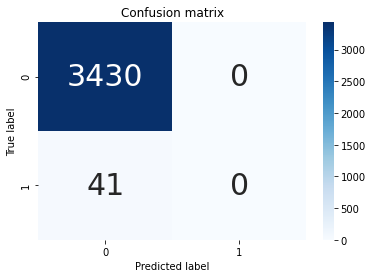}
        \\
        Not SMOTE
    \end{minipage}
    \begin{minipage}{0.32\textwidth} 
        \centering
        \includegraphics[scale=0.33]{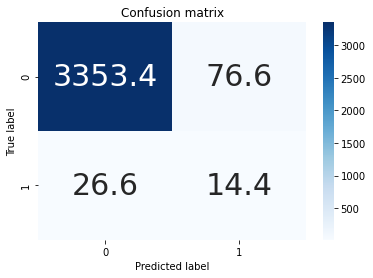}
        \\
        SMOTE
    \end{minipage}
    \caption{Confusion matrix of the Logistic Regression model on the LoL forum.}
    \label{fig:ConfusionmaxtrixSVM}
\end{figure}

For the traditional machine learning model, Figure \ref{fig:ConfusionmaxtrixSVM} shows the confusion matrix of the Logistic Regression model before and after the SMOTE technique is applied. As shown in Figure \ref{fig:ConfusionmaxtrixSVM}, SMOTE increases the predictability of the Logistic Regression model on label 1. However, there are still a lot of comments that are mispredicted from label 1 to label 0.

\begin{figure}[H]
    \begin{minipage}{0.21\textwidth}
        \centering
        \includegraphics[scale=0.33]{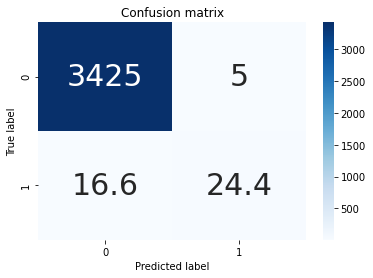}
        \\
        Not SMOTE
    \end{minipage}
    \begin{minipage}{0.32\textwidth}
        \centering
        \includegraphics[scale=0.33]{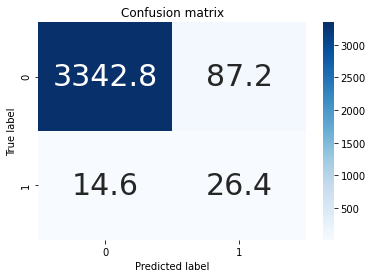}
        \\
        SMOTE
    \end{minipage}
    \caption{Confusion matrix of the Text-CNN with GloVe word embedding on the LoL forum.}
    \label{fig:ConfusionmaxtrixTextCNN_Glove}
\end{figure}

Next, Figure \ref{fig:ConfusionmaxtrixTextCNN_Glove} shows the confusion matrix of the Text-CNN model with the GloVe embedding before and after enhancing the data using the SMOTE technique. After the data enhancement, the amount of data correctly predicted in label 1 is increased by 8.2\%. However, the number of labels 0 correctly predicted decreased by 2.4\%. Therefore, SMOTE does not increase the performance of deep neural models.

\begin{figure}[H]
    \begin{minipage}{0.21\textwidth}
        \centering
        \includegraphics[scale=0.33]{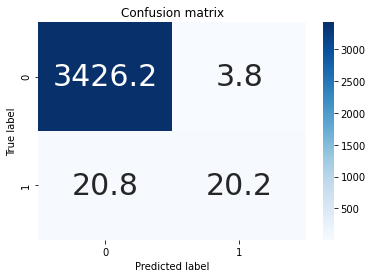}
        \\
        fastText
    \end{minipage}
    \begin{minipage}{0.32\textwidth}
        \centering
        \includegraphics[scale=0.33]{image/lol_textcnn_glove_notsmote.png}
        \\
        GloVe
    \end{minipage}
    \caption{Confusion matrix of the Text-CNN model on the LoL forum (without SMOTE).}
    \label{fig:ConfusionmaxtrixDNN}
\end{figure}

Besides, Figure \ref{fig:ConfusionmaxtrixDNN} shows the confusion matrix of the Text-CNN model on two word embeddings including GloVe and fastText without using SMOTE technique. As shown in Figure \ref{fig:ConfusionmaxtrixDNN}, the predictive accuracy on label 1 of the Text CNN model on GloVe word embedding is better than fastText word embedding. For label 0, the prediction results on both models are similar. Therefore, the performance of the Text-CNN model with GloVe is better than fastText.

\begin{figure}[H]
    \begin{minipage}{0.5\textwidth}
        \centering
        \includegraphics[scale=0.33]{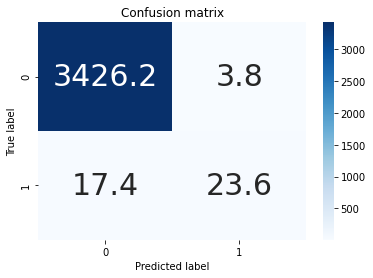}
        \\
    \end{minipage}
    \caption{Confusion matrix of the Toxic-BERT model on the LoL forum.}
    \label{fig:Confusionmaxtrixbert}
\end{figure}

In addition, Figure \ref{fig:Confusionmaxtrixbert} shows the confusion matrix of the Toxic BERT model. According to confusion matrix in Figure \ref{fig:Confusionmaxtrixbert}, the Toxic-BERT model has a higher number of correct predictions on label 1 than the number of incorrect predictions. Comparing with other models, the Toxic-BERT model has a higher total percentage of correct prediction on label 0 and label 1 than other models. Therefore, Toxic-BERT has the highest accuracy on the dataset, even without the data enhancement method. 

Finally, according to Table \ref{tab:vi_du_du_doan_sai}, many comments of players on the forum use acronyms such as "plz" (roughly meaning: please) in comment No \#1, or "u" (roughly meaning: you) in comment No \#4. In addition, the syntax and grammar in player comments are often inaccurate compared to formal texts, for example, the word "previou" (roughly meaning: previous) in comment No \#3, and "minut" (roughly meaning: minute) in comment No \#4. To improve the predictability of the model, it is necessary to apply lexicon based approaches to normalize the comments into formal texts. 

\begin{table}[H]
    \centering
    \caption{Several incorrect predictions in the Cyberbullying dataset.}
    \label{tab:vi_du_du_doan_sai}
    \begin{tabular}{|p{0.3cm}|p{5.2cm}|p{0.6cm}|p{1.1cm}|}
        \hline
        \textbf{No.} & \centering \textbf{Input} &
        \textbf{True} &\textbf{Predicted} \\
        \hline
        1 &  tri argu common sens girl that problem sexist bigot plz never show face realiti kthnx vote oblivion pl x. & 1 & 0 \\
        \hline
        2 & stop give account info idiot absolut true still think account deserv rune back. & 0 & 1 \\
        \hline
        3 &  refund get kill wait oh idiot rescind previou statement trust memori. & 0 & 1 \\
        \hline
        4 & post extra two minut travel much stupid u think min travel. &1 & 0\\
        \hline
        5 &  leav guy troll wont play stupid card fun get attent world wide ignor troll idiot. & 0 & 1\\
        \hline
    \end{tabular}
\end{table}

\section{Conclusion}
\label{conclusion}
In this paper, we introduced a solution to identify offensive comments on game forums by machine learning approaches using text classification models. The Toxic-BERT model got the best results on the Cyberbullying dataset. The weakness of the Cyberbullying dataset is the imbalance between label 1 and label 0, thus leading to too much wrong prediction of label 1. To solve this problem, we used SMOTE for traditional machine learning models and deep neural models to improve the data imbalance, however, results do not improved significantly on deep neural models. From the error analysis, we found that the incorrect predictions were caused by the informal texts used by users on game forums such as acronyms, slangs, and wrong syntax sentences. Therefore, it is necessary to normalize the comments of users to increase the performance of classification models. 

In future, we are going to apply the Easy Data Augmentation techniques (EDA)\cite{wei-zou-2019-eda} to enhance data on minor label, and implement a dictionary to standardize irregular words. The Hurlex \cite{bassignana2018hurtlex} and the abusive lexicon provided by Wiegand et al. \cite{wiegand-etal-2018-inducing} are two potential abusive lexicon sets are used to increase the performance of classification models. Besides, based on the results obtained in this paper, we plan to build a module to automatically detect offensive comments on game forums in order to help moderators for keep the clean and friendly space for discussion among game players.


\bibliographystyle{IEEEtran}
\bibliography{references}

\end{document}